 \documentclass[smallabstract,smallcaptions]{dccpaper}

\usepackage{epsfig}
\usepackage{citesort}
\usepackage{amsmath}
\usepackage{amssymb}
\usepackage{color}
\usepackage{url}
\usepackage{booktabs}

\usepackage{enumitem}
\usepackage{cite}
\usepackage{multirow}
\usepackage{graphicx}
\usepackage{makecell}
\usepackage{pifont}
\newlength{\figurewidth}
\newlength{\smallfigurewidth}

\begin{document}

\title
{\large
% \textbf{L-LBVC: Adaptive Motion Estimation and Prediction for Learned Bi-Directional Video Compression}
% }
\textbf{L-LBVC: Long-Term Motion Estimation and Prediction for Learned Bi-Directional Video Compression}
% }
% \textbf{L-LBVC: Long-Term Motion Estimation for Learned Bi-Directional Video Compression}
}
% and Adaptive Motion Prediction
% Hierarchical

\author{%
Yongqi Zhai$^{1,2}$, Luyang Tang$^{1,2}$, Wei Jiang$^{1}$, Jiayu Yang$^{2}$, Ronggang Wang$^{1,2\ast}$\\[0.5em]
\thanks{$\ast$ Ronggang Wang is the corresponding author.}
{\small\begin{minipage}{\linewidth}\begin{center}
\begin{tabular}{c}
$^{1}$Peking University Shenzhen Graduate School, China \\
$^{2}$Pengcheng Laboratory, China \\
% $^{3}$Migu Culture Technology Co., Ltd, China \\
% Street Address One && Street Address Two \\
% City, State, ZIP, Country && City, State, ZIP, Country\\
\url{zhaiyongqi@stu.pku.edu.cn,}
% \enspace
\thinspace
\url{rgwang@pkusz.edu.cn}
\end{tabular}
\end{center}\end{minipage}}
}

\maketitle
\thispagestyle{empty}

\begin{abstract}
Recently, learned video compression (LVC) has shown superior performance under low-delay configuration. However, the performance of learned bi-directional video compression (LBVC) still lags behind traditional bi-directional coding. The performance gap mainly arises from inaccurate long-term motion estimation and prediction of distant frames, especially in large motion scenes. To solve these two critical problems, this paper proposes a novel LBVC framework, namely L-LBVC. Firstly, we propose an adaptive motion estimation module that can handle both short-term and long-term motions. Specifically, we directly estimate the optical flows for adjacent frames and non-adjacent frames with small motions. For non-adjacent frames with large motions, we recursively accumulate local flows between adjacent frames to estimate long-term flows. Secondly, we propose an adaptive motion prediction module that can largely reduce the bit cost for motion coding. To improve the accuracy of long-term motion prediction, we adaptively downsample reference frames during testing to match the motion ranges observed during training. Experiments show that our L-LBVC significantly outperforms previous state-of-the-art LVC methods and even surpasses VVC (VTM) on some test datasets under random access configuration.

\end{abstract}

% \Section{Introduction}
\Section{Introduction}
With the rapid growth in the amount of video data, it is crucial to design efficient video compression algorithms to reduce storage and transmission costs. In recent years, thanks to the success of deep neural networks, learned video compression has become an active research area. According to different usage scenarios, there are two common configurations: low-delay (LD) and random access (RA).
% Over the past few decades, several traditional video coding standards have been developed, such as H.264/AVC \cite{wiegand2003overview}, H.265/HEVC \cite{sullivan2012overview}, and H.266/VVC \cite{bross2021overview}. However, these methods rely on hand-crafted methods and lack end-to-end optimization, which hinders further performance improvement. 
% With the popularity of video applications in people's daily lives, the amount of video data has grown rapidly. Therefore, it is crucial to design efficient video compression algorithms to reduce storage and transmission costs. 

The LD configuration emphasizes minimizing the coding delay and is suitable for real-time applications, such as video calls and live streaming. Under LD configuration, the predictive frame (P-frame) is encoded sequentially that only refers to previously coded frames. Recently, learned sequential video compression (LSVC) has achieved remarkable progress, and some works \cite{li2022hybrid,zhai2024hybrid,li2023neural,li2024modulation} even outperform H.266/VVC \cite{bross2021overview}. The RA configuration allows the viewer to navigate through a video and is suitable for applications requiring random access, such as video-on-demand. The key frame type of RA configuration is the bi-directional frame (B-frame). Since B-frame takes both past and future frames as references, it theoretically has great potential to achieve a higher compression ratio than P-frame coding, which has been verified in traditional video codecs. However, despite there are some works \cite{yilmaz2021end,yilmaz2023multi,wu2018video,djelouah2019neural,pourreza2021extending,chen2023b,alexandre2023hierarchical,xu2024ibvc,yang2024ucvc} put effort on LBVC, the performance of state-of-the-art (SOTA) LBVC methods still lags behind some LSVC methods \cite{li2023neural,li2024modulation} and the best traditional codec VVC (VTM) under RA configuration.
% The predictive frame (P-frame) is encoded sequentially under the LD configuration, which only refers to previously coded frames to predict the target frame.
% to predict the target frame

% To find the crucial factor causing the performance gap between LBVC and LSVC methods, Fig. \ref{prediction} visualizes the motion estimation and motion compensation (MEMC) results of the SOTA LBVC method B-CANF \cite{chen2023b}. As we can see, the motion estimation method cannot effectively handle large motions between distant frames, leading to inaccurate prediction. 
% LSVC methods have consistent motion ranges between consecutive frames in both training and testing scenarios.
% Chen \textit{et al.} \cite{chen2023b} noted that the root cause of reducing the motion estimation accuracy of LBVC is the domain shift between the training and test scenarios. 
% Chen \textit{et al.} \cite{chen2023b} noted that the domain shift between the training and test scenarios reduces the motion estimation accuracy of LBVC.
% A few works have attempted to solve this problem. Xu \textit{et al.} \cite{xu2024ibvc} periodically inserted P-frames to make the temporal prediction distance constant and reduce motion diversity. Yang \textit{et al.} \cite{yang2024ucvc} selected B-frames for slow motion scenes and P-frames for large motion scenes.
There are two critical problems that cause the low compression performance of existing LBVC methods. The first problem is the inaccurate long-term motion estimation between distant frames, especially in large motion scenes. Existing LBVC methods \cite{chen2023b,yilmaz2021end,yilmaz2023multi,alexandre2023hierarchical} are trained on a small group of pictures (GoP) with small motion ranges, but tested on large GoPs with various motion ranges. The domain shift between training and testing scenarios \cite{chen2023b} results in the inability of the motion estimation module to handle long-term motions. Hence, existing LVC methods perform well under LD configuration but perform poorly under RA configuration. Although some works \cite{xu2024ibvc,yang2024ucvc} attempt to solve this problem, the accuracy of long-term motion estimation is not fundamentally improved. To this end, we propose an adaptive motion estimation (AME) module, which applies different estimation methods for different types of motion. Specifically, we directly estimate the optical flows for adjacent frames and non-adjacent frames with small motions. For long-term large motions between non-adjacent frames, simply using the direct estimation method will cause the accuracy of motion estimation to be severely degraded or even become unacceptable. Thus, to obtain accurate long-term motions, we recursively accumulate local flows between adjacent frames and design an optical flow refinement module (OFR) to rectify the accumulation error. In this way, our proposed AME module can effectively estimate both short-term and long-term motions. 

The second problem is that the motion prediction accuracy of existing LBVC methods is still insufficient. Since compressing bi-directional motions requires more bit cost, how to accurately predict bi-directional motions using forward and backward reference frames is vital for improving the motion compression performance. Many existing works \cite{chen2023b,yilmaz2023multi,alexandre2023hierarchical} utilize video frame interpolation techniques for motion prediction. However, due to the domain shift problem, the accuracy of long-term large motion prediction becomes unacceptable. In this paper, we propose an adaptive prediction (AMP) module that can largely reduce the bit cost for motion coding. During testing, the AMP module selects the optimal spatial resolution of reference frames before conducting flow prediction, thereby adjusting the motion range to the scale observed during training. In addition, when there are a large number of errors in the motion prediction results, the AMP module decides not to use the motion prediction results to avoid a negative impact on motion coding. 

Our main contributions are summarized as follows:

% With these effective techniques, we propose a novel LBVC framework, namely L-LBVC.
% The second problem is that compressing bi-directional motions requires more bit cost. So how to accurately predict bi-directional motions using the forward and backward reference frames is vital for reducing the bit cost for motion coding. Many existing works \cite{chen2023b,yilmaz2023multi,alexandre2023hierarchical} utilize video frame interpolation technique to perform motion prediction. However, due to the domain shift problem, the motion prediction accuracy is still insufficient and even becomes unacceptable for long-term motions. In this paper, we propose an adaptive prediction (AMP) module to reduce the bit cost for motion coding. During testing, the AMP module selects the optimal spatial resolution of reference frames before conducting flow prediction to bring the motion range to the scale observed during training. Moreover, when there are a large number of errors in the motion prediction results, the AMP module decides not to use the motion prediction results to avoid a negative impact on the performance of motion coding. With these effective techniques, we propose a novel LBVC framework, namely L-LBVC. Our main contributions are summarized as follows:

% With these effective techniques, we propose a novel framework L-LBVC, which significantly improves the rate-distortion (RD) performance of LBVC. Our main contributions are summarized as follows:

\begin{itemize}[topsep=0pt, itemsep=2pt, parsep=1pt]
\item[\small\textbullet] We propose an adaptive motion estimation (AME) module to handle both short-term and long-term motions. The proposed accumulation-based estimation method significantly improves the accuracy of long-term motion estimation.
% \item[\small\textbullet] We propose an adaptive motion estimation (AME) module to improve the accuracy of long-term motion estimation, which can handle various motion ranges.
% and obtain accurate prediction results
\item[\small\textbullet] We propose an adaptive motion prediction (AMP) module to reduce the bit cost for motion coding, which selects the best resolution of reference frames for motion prediction and decides whether to use the motion prediction results. 

% which selects the best resolution of reference frames for motion prediction to 
% adaptively downsamples the reference frames
% can improve the improve the accuracy of long-term motion prediction.

% The proposed AMP method not only selects the best resolution of reference frames for motion prediction, but also decides whether to use motion prediction results.
\item[\small\textbullet] Experimental results show that in terms of PSNR, our proposed L-LBVC significantly exceeds previous SOTA LVC methods and even achieves better performance than VVC (VTM) on some test datasets under RA configuration.
% and comparable performance to VVC under random access configuration in terms of PSNR.
\end{itemize}
% it will have a negative impact on the compression performance of the motion. In this case, the AMP will decides not to use the motion prediction to
% the AMP decides not to use the motion prediction to 
% using the predicted optical flow has a negative impact on compression performance.
% to suit the capability of the motion estimation bring motion ranges to a scale observed in the training data and mitigate data drift during flow prediction stage.
% Our proposed adaptive motion prediction method selects the best resolution of reference frames for optical flow prediction
% Furthermore, the compression efficiency of B-frame may be worse than intra frame (I-frame) when a a frame suffers from large occlusion \cite{brand2024conditional,chen2024maskcrt}. We introduce a conditional intra frame (cI-frame) for such scenarios. The cI-frame reduces temporal redundancy through an adaptive spatial-temporal entropy model and achieves higher compression efficiency than B-frame and I-frame. To boost LBVC, we also improve two key techniques of LBVC, namely adaptive optical flow prediction and bi-directional context enhancement. The adaptive optical flow prediction method not only decides whether to use optical flow prediction, but also selects the best resolution of reference frames for flow prediction. The bi-directional context enhancement method adopts multi-scale deformable alignment, which improves the quality of contexts and reduces the bit-rate for contextual coding. 

\begin{figure*}[!t]
% \vspace{-1cm}
  \centering
  \includegraphics[width=\linewidth, trim = {0 0 0 0}, clip]{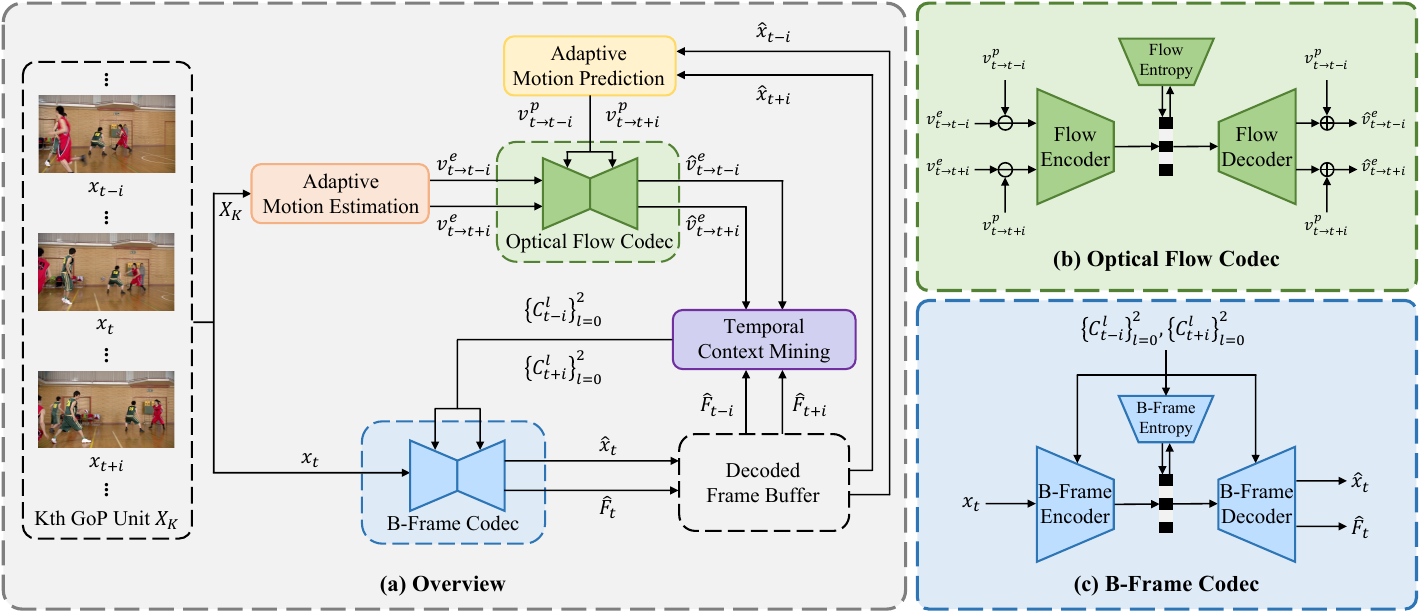}
  % \vspace{-0.5cm}
  \caption{(a) Overview of the proposed L-LBVC framework. (b) Illustration of the optical flow codec. (c) Illustration of the B-frame codec.}
  \label{overall_framework}
\vspace{-0.3cm}
\end{figure*}

\Section{Proposed Method}
\SubSection{Overview}
Let $X=\{x_0, x_1, \cdots, x_t, \cdots\}$ denotes the original video sequence, where $x_t$ represent the frame at time step $t$. We divide $X$ into several GoPs and compress them separately. 
% For I-frame compression, we use the image codec released in DCVC-DC \cite{li2023neural}. For B-frame compression, as shown in Fig.~\ref{overall_framework}, our L-LBVC adopts the conditional coding framework \cite{li2023neural} and consists of the following main components:
% Fig.~\ref{overall_framework} (a) shows the overview of our L-LBVC framework.
% We take the DCVC-DC \cite{li2023neural} as the backbone but do not apply the offset diversity module to reduce complexity. 
As shown in Fig.~\ref{overall_framework}, our L-LBVC is based on DCVC-DC \cite{li2023neural} (without offset diversity module), which consists of the following main components:
% Our L-LBVC is build on DCVC-DC \cite{li2023neural} but not apply the the offset diversity module to reduce complexity.

\textbf{Adaptive Motion Estimation (AME).} The AME module takes all the original images of the Kth GoP unit $X_{K}$ as input and estimates the bi-directional optical flows $v^e_{t\rightarrow{t-i}}$, $v^e_{t\rightarrow{t+i}}$ between the current frame $x_t$ and the two original reference frames $x_{t-i}$, $x_{t+i}$. As for optical flow estimation, we use SpyNet \cite{ranjan2017optical} for acceleration.

\textbf{Adaptive Motion Prediction (AMP).} Given the two decoded reference frames $\hat{x}_{t-i}$, $\hat{x}_{t+i}$, the AMP module estimates two optical flows $v^p_{t\rightarrow{t-i}}$, $v^p_{t\rightarrow{t+i}}$ as the predictions of $v^e_{t\rightarrow{t-i}}$, $v^e_{t\rightarrow{t+i}}$. We use RIFE \cite{huang2022real} as our optical flow predictor.

\textbf{Optical Flow Codec.} We calculate the motion differences $v^r_{t\rightarrow{t-i}}$, $v^r_{t\rightarrow{t+i}}$ between the original motions $v^e_{t\rightarrow{t-i}}$, $v^e_{t\rightarrow{t+i}}$ and the predicted motions $v^p_{t\rightarrow{t-i}}$, $v^p_{t\rightarrow{t+i}}$. Then we encode $v^r_{t\rightarrow{t-i}}$, $v^r_{t\rightarrow{t+i}}$ jointly to obtain the reconstructed motions $\hat{v}^e_{t\rightarrow{t-i}}$, $\hat{v}^e_{t\rightarrow{t+i}}$. 

% The differences between the original motions and the predicted motions are 

\textbf{Temporal Context Mining.} Based on the decoded motions $\hat{v}^e_{t\rightarrow{t-i}}$, $\hat{v}^e_{t\rightarrow{t+i}}$, we warp the reference features $\hat{F}_{t-i}$, $\hat{F}_{t+i}$ instead of decoded frames $\hat{x}_{t-i}$, $\hat{x}_{t+i}$ to generate the multi-scale temporal contexts $\{{C}_{t-i}^l\}_{l=0}^2$, $\{{C}_{t+i}^l\}_{l=0}^2$.
% Like the P-frame conditional coding frameworks \cite{sheng2022temporal,li2022hybrid}, 
% we generate multi-scale temporal contexts $\{{C}_{t-i}^l\}_{l=0}^2$, $\{{C}_{t+i}^l\}_{l=0}^2$ from reference features $\hat{F}_{t-i}$, $\hat{F}_{t+i}$ instead of decoded frames $\hat{x}_{t-i}$, $\hat{x}_{t+i}$.

\textbf{B-Frame Codec.} Conditioned on the multi-scale temporal contexts, the current frame $x_t$ is compressed by the B-frame codec and decoded as intermediate feature $\hat{F}_{t}$ and reconstructed frame $\hat{x}_{t}$. $\hat{F}_{t}$ is referenced when encoding/decoding the next frame.
% the B-frame codec encodes and decodes the current frame $x_t$. At the decoder side, the B-frame decoder decodes high-resolution feature $\hat{F}_{t}$ and the reconstructed frame $\hat{x}_{t}$.
% and generates reconstructed frame and feature.
% the current frame $x_t$ is compressed by the B-frame encoder÷ At the decoder side, the 
% 采用条件编码框架，包括几个部分
\begin{figure*}[!t]
% \vspace{-1cm}
  \centering
  \includegraphics[width=\linewidth, trim = {0 0 0 0}, clip]{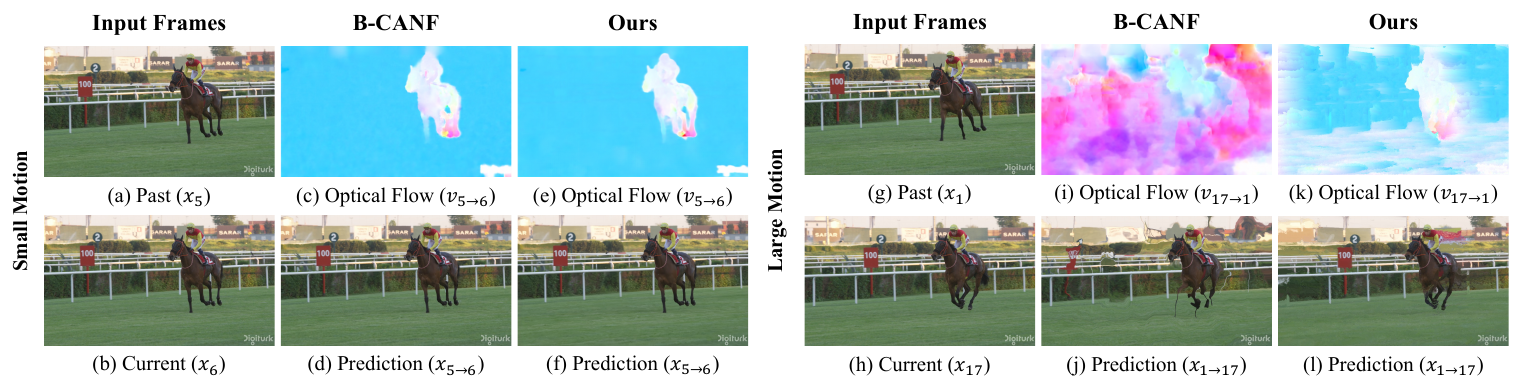}
  \vspace{-0.9cm}
  \caption{Comparisons of the estimated optical flows and temporal predictions of our L-LBVC with B-CANF \cite{chen2023b} on \textit{Jockey} dataset.}
  \label{view_accflow}
\vspace{-0.4cm}
\end{figure*}
\vspace{-0.6cm}

\SubSection{Adaptive Motion Estimation}
To find the crucial factor causing the low compression performance of existing LBVC methods, we visualize the motion estimation and temporal prediction results of the SOTA LBVC method B-CANF \cite{chen2023b} in Fig.~\ref{view_accflow}. As we can see, the motion estimation module of B-CANF performs well in small motion scenes. However, when the frame distance increases, B-CANF cannot effectively handle the large motions between distant frames, leading to inaccurate prediction. Unlike LSVC, which has a uniform motion range between adjacent frames, existing LBVC methods \cite{chen2023b,yilmaz2021end,yilmaz2023multi,alexandre2023hierarchical} cannot adapt to various motion ranges, resulting in performance loss. 

% Chen \textit{et al.} \cite{chen2023b} noted that the root cause of the low motion estimation accuracy of LBVC is the domain shift between the training and test scenarios.
% So the key is how to design an effective motion estimation method to handle various motion ranges. 
% We adaptively select the best motion estimation method according to the motion intensity of two frames. 
In this paper, we propose an adaptive motion estimation (AME) module to address the above problem. Our AME module consists of two optical flow estimation methods: direct estimation and accumulation-based estimation. The direct estimation method directly inputs two frames ${x}_{t}$, ${x}_{t-i}$ into SpyNet \cite{ranjan2017optical} to estimate optical flow $v^e_{t\rightarrow{t-i}}$, which performs well in small motion scenes. However, as the time interval increases, the accuracy of optical flow estimation severely decreases or even becomes unacceptable due to the domain shift problem.

\begin{figure*}[!t]
% \vspace{-1cm}
  \centering
  \includegraphics[width=\linewidth, trim = {0 0 0 0}, clip]{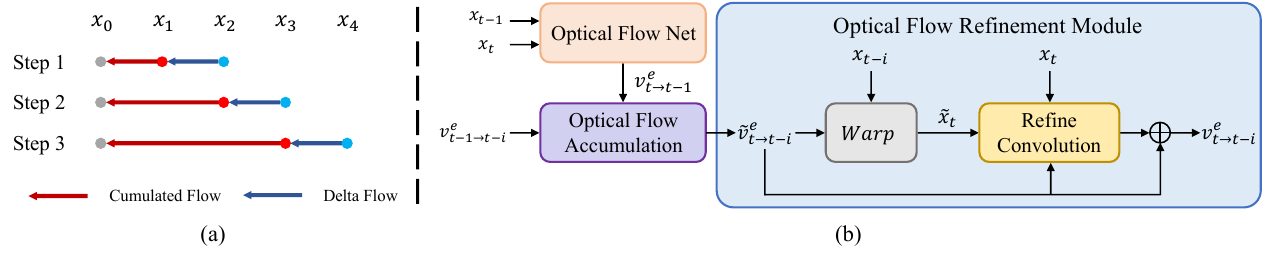}
  \vspace{-1cm}
  \caption{Illustration of the proposed accumulation-based optical flow estimation method. (a) Example of the optical flow accumulation process. (b) The detailed network structure.}
  \label{accflow}
\vspace{-0.5cm}
\end{figure*}

% Therefore, we apply direct optical flow estimation for adjacent frames and non-adjacent frames with small motions.
To accurately estimate long-term motions, we propose to recursively accumulate local flows between adjacent frames. Fig.~\ref{accflow} (a) provides an example of the optical flow accumulation process for estimating flow $v_{4\rightarrow0}$ between $x_4$ and $x_0$. We first estimate local optical flows $\{v_{1\rightarrow0}$, $v_{2\rightarrow1}$, $v_{3\rightarrow2}$, $v_{4\rightarrow3}\}$ between original adjacent frames. In step 1, to obtain the flow $v_{2\rightarrow0}$ between $x_2$ and $x_0$, we warp $v_{1\rightarrow0}$ towards $v_{2\rightarrow1}$ via:
\begin{equation}
    v_{2\rightarrow0} = \mathcal{W}(v_{1\rightarrow0}, v_{2\rightarrow1}) + v_{2\rightarrow1},
\end{equation}
where $\mathcal{W}$ denotes the warping operator. In step 2 and step 3, the intermediate local optical flows are recursively accumulated to the target optical flow $v_{4\rightarrow0}$ via:
% \vspace{-0.15cm}
% \begin{equation}
%     v_{3\rightarrow0} = \mathcal{W}(v_{2\rightarrow0}, v_{3\rightarrow2}) + v_{3\rightarrow2}, 
% \end{equation}
% % \vspace{-0.6cm}
% \begin{equation} 
%     v_{4\rightarrow0} = \mathcal{W}(v_{3\rightarrow0}, v_{4\rightarrow3}) + v_{4\rightarrow3}.
% \end{equation}
\begin{equation}
\begin{split}
    v_{3\rightarrow0} = \mathcal{W}(v_{2\rightarrow0}, v_{3\rightarrow2}) + v_{3\rightarrow2}, 
    \\
    v_{4\rightarrow0} = \mathcal{W}(v_{3\rightarrow0}, v_{4\rightarrow3}) + v_{4\rightarrow3}.
\end{split}
\end{equation}
Due to the influence of occlusion, the optical flow accumulation process will bring the accumulation error, resulting in inaccurate optical flow estimation in non-occluded areas. To tackle this critical issue, we design an optical flow refinement (OFR) module to rectify the error in each accumulation process. The detailed network structure of the optical flow refinement module is shown in Fig.~\ref{accflow} (b). Based on the accumulated flow $\tilde{v}^e_{t\rightarrow{t-i}}$, we first warp $x_{t-i}$ to obtain the prediction frame $\tilde{x}_t$ of the current frame $x_t$. Then, we design a refine convolution block which consists of several $3\times3$ convolution layers to refine the initial accumulated flow:
\begin{equation}
    {v}^e_{t\rightarrow{t-i}} = \mathit{Conv}(x_t, \tilde{x}_t, \tilde{v}^e_{t\rightarrow{t-i}}) + \tilde{v}^e_{t\rightarrow{t-i}},
\end{equation}
where $\mathit{Conv}$ represents several convolution layers.

Meanwhile, we evaluate the motion intensity between two frames with motion intensity factor (MIF), which is calculated by:
\begin{equation}
    MIF = \frac{\sum_{i=0}^{H-1}\sum_{j=0}^{W-1}\sqrt{(m^{i,j}_x)^2 + (m^{i,j}_y)^2}}{HW},
\end{equation}
where $m_x$ and $m_y$ denote horizontal and vertical offsets of accumulated flow, H and W denote the height and width of the frame. If the average MIF of the bi-directional optical flows is less than a given threshold T, we define it as small motion and use the direct optical flow estimation method. Otherwise, we define it as large motion and use the accumulation-based optical flow estimation method. The default threshold T is set as 10. As shown in Fig.~\ref{view_accflow}, our proposed AME module achieves accurate motion estimation and temporal prediction in both small and large motion scenes.
% Specifically, to obtain the non-adjacent optical flow $v^e_{t\rightarrow{t-i}}$

% Our AME method adaptively selects the best motion estimation method among the following methods : direct optical flow estimation method and accumulation-based optical flow estimation method.
% if directly estimating optical flow in large motion scenes, the accuracy of optical flow will severely decrease or even become unacceptable.
\begin{figure*}[!t]
% \vspace{-1cm}
  \centering
  \includegraphics[width=\linewidth, trim = {0 0 0 0}, clip]{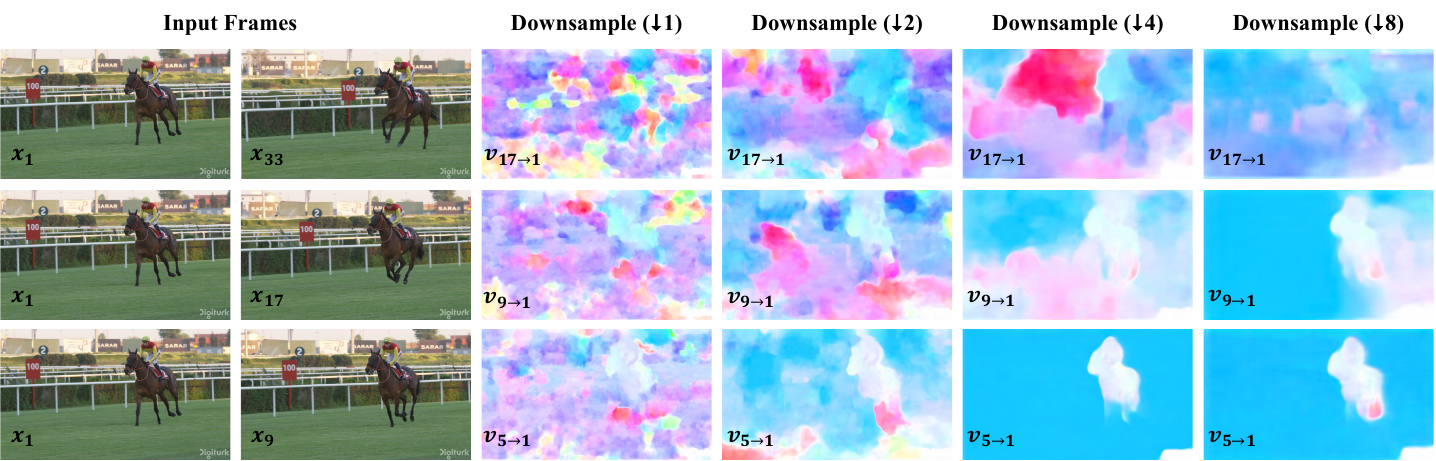}
  % \vspace{-1cm}
  \caption{Visualization of the predicted optical flows using different downsampling factors.}
  \label{amp}
\vspace{-0.3cm}
\end{figure*}

% \vspace{-0.3cm}
% As we can see, the motion estimation method cannot effectively handle large motions between distant frames, leading to inaccurate prediction. Chen \textit{et al.} \cite{chen2023b} noted that the root cause of reducing the motion estimation accuracy of LBVC is the domain shift between the training and test scenarios. 
\SubSection{Adaptive Motion Prediction}
Compared with LSVC, LBVC needs to compress bi-directional motions to utilize the forward and backward reference frames, thus requiring more motion coding costs. Although some works \cite{chen2023b,yilmaz2023multi,alexandre2023hierarchical} design motion prediction modules to predict bi-directional motions, as shown in the third column of Fig.~\ref{amp}, these modules suffer from the domain shift and perform poorly when predicting long-term motions. 

To make the optical flow predictor handle consistent motion range between training and testing scenarios, we propose an adaptive motion prediction (AMP) module that adapts the motion range by adaptively downsampling the reference frames for flow prediction during testing. It is simple but quite effective. At the encoder side, we select the downsampling factor $S{\in}\{1,2,3,4\}$ that achieves the best prediction performance. Specifically, we first downsample the forward and backward reference frames $\hat{x}_{t-i}$, $\hat{x}_{t+i}$ by a factor $S$ and then input them into optical flow predictor RIFE \cite{huang2022real}:
\begin{equation}
    v^{p(\downarrow{S})}_{t\rightarrow{t-i}}, v^{p(\downarrow{S})}_{t\rightarrow{t+i}} = RIFE(\hat{x}^{(\downarrow{S})}_{t-i}, \hat{x}^{(\downarrow{S})}_{t+i}).
\end{equation}
We upsample the predicted flows to the original resolution $v^{p({S})}_{t\rightarrow{t-i}}$, $v^{p({S})}_{t\rightarrow{t+i}}$ and then warp the reference frames $\hat{x}_{t-i}$, $\hat{x}_{t+i}$ to get the predicted frames $\bar{X}^{(S)}_{t-i}$, $\bar{X}^{(S)}_{t+i}$.
% based on $v^{p({S})}_{t\rightarrow{t-i}}$, $v^{p({S})}_{t\rightarrow{t+i}}$
% \begin{equation}
% \begin{split}
%     \bar{X}^{(S)}_{t-i} = \mathcal{W}(\hat{x}_{t-i}, v^{p({S})}_{t\rightarrow{t-i}}), 
%     \\
%     \bar{X}^{(S)}_{t+i} = \mathcal{W}(\hat{x}_{t+i}, v^{p({S})}_{t\rightarrow{t+i}}).
% \end{split}
% \end{equation}
We take the average of the two predicted frames $\bar{X}^{(S)}_{t-i}$, $\bar{X}^{(S)}_{t+i}$ as the final predicted frame and calculate its $PSNR(S)$:
\begin{equation}
    PSNR(S) = psnr(x_t, 0.5(\bar{X}^{(S)}_{t-i} + \bar{X}^{(S)}_{t+i})).
\end{equation}
The factor $S_{opt}$ that obtains the best PSNR value is chosen. 

As shown in the last two rows of Fig.~\ref{amp}, we successfully predict the long-term optical flow by downsampling the reference frames before conducting flow prediction. However, when the optimal motion prediction result still has a lot of errors (see the first row of Fig.~\ref{amp}), it will have a negative impact on motion coding. When $PSNR(S)$ is smaller than a given threshold P, we set the predicted optical flows to 0, which means do not use the prediction results. In our implementation, the threshold P is experimentally set to 20. Finally, the best downsampling factor $S$ and the flag on whether to use the prediction results are signaled in the bitstream.
% \SubSection{Hierarchical Quality Structure}
% Inspired by the success of hierarchical quality structure (HQS) in traditional codec, we propose to assign different distortion weights $w_t$ for different temporal layers during training. The RD loss function is defined as: 
% \begin{equation}
%     \mathcal{L} =R + w_t \cdot \lambda \cdot D=R_{\hat v}+R_{\hat f}+ w_t \cdot \lambda \cdot D(x_t, \hat{x}_t).
% \end{equation}
% $R_{\hat v}$ and $R_{\hat f}$ respectively denote the bitrate of the optical flow coding and the frame coding. $D(\cdot)$ denotes the distortion, which can be $L_2$ loss or MS-SSIM. The $w_t$ settings of B-frames for 5 temporal levels (GoP size 32) are (1.1, 1.1, 0.7, 0.6, 0.5). 
% Given the forward and backward reference frames $\hat{x}_{t-i}$, $\hat{x}_{t+i}$,
% 传输下采样倍数
% these methods suffer from the domain shift problem when predicting long-term motions. As shown in the third column of Fig.~\ref{view_accflow}, the optical flow predictor RIFE \cite{huang2022real} used in \cite{chen2023b, alexandre2023hierarchical} performs poorly in large motion scenes. 
% \vspace{-0.3cm}
% \SubSection{Hierarchical Quality Structure}
\Section{Experiments}
\label{sec:experiments}
\SubSection{Experimental Setup}
\label{sec:experimental_setup}
\textbf{Datasets.}
We train our models on 256×256 random crops of the Vimeo-90k \cite{xue2019video} dataset. For testing, we use the UVG \cite{mercat2020uvg}, MCL-JCV \cite{wang2016mcl} and HEVC \cite{sullivan2012overview} datasets. These datasets contain videos of different motion intensities and resolutions, which are widely used to evaluate the performance of LVC methods.

\noindent
% \textbf{Implementation Details.}
\textbf{Training Details.}
% Inspired by the success of hierarchical quality structure in traditional codec, we propose to assign different distortion weights $w_t$ for different temporal layers during training. The RD loss function is defined as: $\mathcal{L} =R + w_t \cdot \lambda \cdot D=R_{\hat v}+R_{\hat f}+ w_t \cdot \lambda \cdot D(x_t, \hat{x}_t)$. $R_{\hat v}$ and $R_{\hat f}$ respectively denote the bitrate of the optical flow coding and the frame coding. $D(\cdot)$ denotes the distortion, which can  be $L_2$ loss or MS-SSIM. The $w_t$ settings of B-frames for  5 temporal levels (GoP size 32) are (1.1, 1.1, 0.7, 0.6, 0.5). 
We use the multi-stage training strategy \cite{li2023neural} and gradually increase the number of B-frames from 1 to 5 frames. The RD loss function is defined as: $\mathcal{L} =R + w_t \cdot \lambda \cdot D=R_{\hat v}+R_{\hat f}+ w_t \cdot \lambda \cdot D(x_t, \hat{x}_t)$. $R_{\hat v}$ and $R_{\hat f}$ respectively denote the bitrate of the optical flow coding and the frame coding. $D(\cdot)$ denotes the distortion (MSE). The distortion weights $w_t$ settings of B-frames for 5 temporal levels (GoP size 32) are (1.1, 1.1, 0.7, 0.6, 0.5). We set 4 ${\lambda}$ values (256, 512, 1024, 2048) to control the RD trade-off. We employ the AdamW optimizer and set the batch size as 4.

% The RD loss function is defined as: $\mathcal{L} =R + \lambda D=R_{\hat v}+R_{\hat f}+ \lambda D(x_t, \hat{x}_t)$. $R_{\hat v}$ and $R_{\hat f}$ respectively denote the bitrate of the optical flow coding and the frame coding. $D(\cdot)$ denotes the distortion (MSE).

\begin{figure}[!t]
  \centering
  % \vspace{-0.2cm}
  \includegraphics[width=\linewidth, trim = {10 0 10 10}, clip]{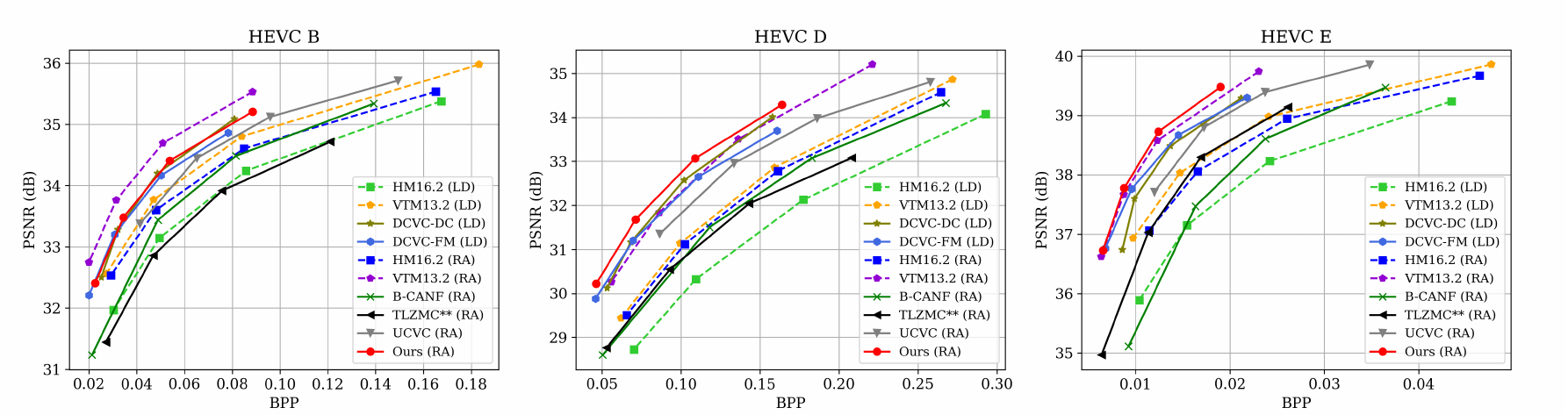}
  \vspace{-1.3cm}
  \caption{RD-curves on the HEVC Class B, D and E datasets.}
  \label{rd_curves}
  \vspace{-0.5cm}
\end{figure}

\begin{table}[!t]
\renewcommand{\arraystretch}{1.1}
\scriptsize
\centering
\vspace{-0.3cm}
\caption{BD-Rate (\%) comparison for PSNR. The anchor is HM-16.20\,(RA).}\label{bdrate_psnr}
\vspace{-0.3cm}
\resizebox{0.95\textwidth}{!}{
% \fontsize{8}{15}\selectfont%设置字体大小 %\midrule
\begin{tabular}{l|cccccc|c}
    \toprule  
        \makebox[0.12\textwidth][c]{Methods} & \makebox[0.06\textwidth][c]{HEVC B} & \makebox[0.06\textwidth][c]{HEVC C} & \makebox[0.06\textwidth][c]{HEVC D} & \makebox[0.06\textwidth][c]{HEVC E} & \makebox[0.01\textwidth][c]{UVG} & \makebox[0.07\textwidth][c]{MCL-JCV} & \makebox[0.03\textwidth][c]{Avg} \\
    \hline  
    % HM-16.20  &0.0 &0.0 &0.0 &0.0 &0.0 &0.0 &0.0 &0.0 \\ %\midrule
    HM-16.20\,(LD)  &25.4 &26.7 &31.3 &34.6 &27.2 &24.6 &28.3 \\ 
    VTM-13.2\,(LD)  &-11.4 &-9.0 &-3.6 &-10.4 &-6.5 &-13.1 &-9.0 \\ %\midrule
    DCVC-DC\,(LD)  &-22.0 &-19.9 &-32.3 &-24.6 &-25.0 &-22.8 &-24.4 \\ 
    %\midrule\midrule
    DCVC-FM\,(LD)  &-20.5 &-19.3 &-31.1 &-30.5 &-23.7 &-20.7 &-24.3 \\
    \hline 
    HM-16.20\,(RA)  &0.0 &0.0 &0.0 &0.0 &0.0 &0.0 &0.0 \\ 
    VTM-13.2\,(RA) &\textbf{-41.4} &\textbf{-33.5} &-30.7 &-42.0 &\textbf{-40.9} &\textbf{-38.6} &\textbf{-37.8} \\
    B-CANF\,(RA) &7.0 &27.6 &4.3 &13.6 &6.2 &12.3 &11.8 \\
    TLZMC**\,(RA) &35.4 &38.3 &8.2 &-6.9 &17.1 &34.3 &21.0 \\
    UCVC\,(RA) &-14.8 &-2.4 &-19.4 &-26.7 &-15.0 &1.9 &-12.7 \\
    Ours\,(RA)  &-25.7 &-14.1 &\textbf{-38.8} &\textbf{-43.6} &-29.0 &-15.6 &-27.8 \\ %\midrule
    \bottomrule 
\end{tabular}
}
% \vspace{-0.5cm}
\end{table}

\begin{table}[!t]
\renewcommand{\arraystretch}{1.1}
\scriptsize
\centering
\vspace{-0.3cm}
 \caption{Ablation study on proposed methods.}\label{ablation_study_overall}
\vspace{-0.3cm}
\resizebox{0.95\textwidth}{!}{
% \fontsize{7}{10}\selectfont%设置字体大小 %\midrule
\begin{tabular}{ccc|cccccc|c}
    \toprule  
        \makebox[0.06\textwidth][c]{Methods} & \makebox[0.03\textwidth][c]{AME} & \makebox[0.03\textwidth][c]{AMP} & \makebox[0.06\textwidth][c]{HEVC B} & \makebox[0.06\textwidth][c]{HEVC C} & \makebox[0.06\textwidth][c]{HEVC D} & \makebox[0.06\textwidth][c]{HEVC E} & \makebox[0.03\textwidth][c]{UVG} & \makebox[0.08\textwidth][c]{MCL-JCV} & \makebox[0.03\textwidth][c]{Avg} \\
    \hline  % &\textbf{\makecell{Flow-based warping \\ at all scale}} \makebox[0.05\textwidth][c]
    A  &\ding{55} &\ding{55} &0.0 &0.0 &0.0 &0.0 &0.0 &0.0 &0.0 \\ 
    B   &\ding{52} &\ding{55} &-10.6 &-3.3 &-5.8 &0.0 &-18.7 &-9.0 &-7.9 \\ %\midrule
    C   &\ding{52} &\ding{52} &\textbf{-20.9} &\textbf{-7.9} &\textbf{-15.6} &\textbf{-6.5} &\textbf{-29.9} &\textbf{-14.2} &\textbf{-15.8} \\ %\midrule\midrule
    \bottomrule 
\end{tabular}
}
\vspace{-0.2cm}
\end{table}

% \vspace{-0.4cm}
\SubSection{Experimental Results}
\label{sec:experimental_results}
\begin{sloppypar}
\textbf{Settings of Competitors.}
To compare with traditional codecs, we test the best encoder of HEVC (HM-16.20) and VVC (VTM-13.2) with default configurations. As for learned video compression methods, we compare with SOTA sequential compression methods DCVC-DC \cite{li2023neural}, DCVC-FM \cite{li2024modulation} and SOTA bi-directional compression methods B-CANF \cite{chen2023b}, TLZMC**  \cite{alexandre2023hierarchical}, UCVC \cite{yang2024ucvc}. To make a fair comparison, we set the intra period to 32 for all schemes and test 96 frames for each video.
% For HM-16.20, we use the default \textit{encoder\_lowdelay\_main.cfg} and \textit{encoder\_randomaccess\_main.cfg} configurations. For VTM-13.2, we use the default \textit{encoder\_lowdelay\_vtm.cfg} and \textit{encoder\_randomaccess\_vtm.cfg} configurations.
\end{sloppypar}

\noindent
\textbf{Results.} Tab~\ref{bdrate_psnr} shows the BD-Rate (\%) comparisons in terms of PSNR. We also draw the RD-curves on the HEVC B, D, and E datasets in Fig.~\ref{rd_curves}. As we can see, our proposed L-LBVC obtains average bitrate savings of 27.8$\%$ against the anchor HM-16.20 (RA) on all test datasets, which outperforms the SOTA LBVC methods B-CANF (11.8\%), TLZMC (21.0\%) and UCVC (-12.7\%). On the HEVC Class D and E datasets, our L-LBVC even surpasses VTM-13.2 (RA). In addition, our L-LBVC achieves better compression performance compared with the SOTA LSVC methods DCVC-DC (-24.4\%) and DCVC-FM (-24.3\%). Note that to reduce complexity and speed up training, our L-LBVC does not apply the offset diversity module that has been shown to be effective in DCVC-DC and DCVC-FM.

% the compression performance of our L-LBVC will be better if we apply the group-based offset diversity module, which has been shown to be effective in DCVC-DC and DCVC-FM.

% Note that to reduce complexity, we do not apply the group-based offset diversity module that has been shown to be effective in DCVC-DC and DCVC-FM. The compression performance of our L-LBVC will be better if we apply this module.

\SubSection{Ablation Study}
\textbf{Effectiveness of Proposed Modules.}
We conduct ablation studies in Tab~\ref{ablation_study_overall} to verify the effectiveness of our proposed modules. Method A adopts direct flow estimation and is set as anchor. We progressively add our proposed adaptive motion estimation (AME) module and adaptive motion prediction (AMP) module to method A. By comparing methods A and B, we find our AME module achieves average bitrate savings of -7.9\%, which indicates the effectiveness of the AME module. Since the AME module adopts direct estimation for small motions, there is no gain is achieved on the HEVC E dataset that only contains small motions. Our full model (method C) achieves better RD performance than method B on all test datasets, which demonstrates that the AMP module can largely reduce the bit cost for motion coding. 

\begin{table}[!t]
\centering
\renewcommand{\arraystretch}{1.1}
\scriptsize
% \vspace{-0.5cm}
\caption{Ablation study on adaptive motion estimation method.}\label{ablation_study_ame}
\vspace{-0.3cm}
\resizebox{0.98\textwidth}{!}{
% \fontsize{7}{10}\selectfont%设置字体大小 %\midrule
% Accumulation-based Estimation (/wo OFR)
% Accumulation-based Estimation (/w OFR)
\begin{tabular}{c|c|cccc}
    \toprule  
        % \makebox[0.06\textwidth][c]{Sequences} & \makebox[0.06\textwidth][c]{Deirect} & \makebox[0.18\textwidth][c]{Accumulation\,(/wo OFR)} & \makebox[0.18\textwidth][c]{Accumulation\,(/w OFR)} & \makebox[0.06\textwidth][c]{Adaptive}\\
        \multirow{2}{*}{\makebox[0.06\textwidth][c]{Motion Type}} & \multirow{2}{*}{\makebox[0.05\textwidth][c]{Sequences}} & \multirow{2}{*}{\makebox[0.05\textwidth][c]{Deirect}} & \makebox[0.11\textwidth][c]{Accumulation} & \makebox[0.11\textwidth][c]{Accumulation} & \multirow{2}{*}{\makebox[0.03\textwidth][c]{AME}}\\
        & & & (w/o OFR) & (w/ OFR) &  \\
        
    \hline  % &\textbf{\makecell{Flow-based warping \\ at all scale}} \makebox[0.05\textwidth][c]
    & BasketballDrive   &0.0 &-20.0 &-31.5 &\textbf{-31.5} \\
    & Kimono   &0.0 &-14.6 &-27.9 &\textbf{-27.9} \\
    Large Motion & Jockey   &0.0 &-30.6 &-50.4 &\textbf{-50.4} \\ %\midrule
    & ReadySteadyGo   &0.0 &-17.4 &-51.2 &\textbf{-51.2} \\ %\midrule
    & YatchRide   &0.0 &-9.7 &-25.4 &\textbf{-25.4} \\ %\midrule
    \hline
    & HoneyBee   &0.0 &19.8 &13.1 &\textbf{0.0} \\ %\midrule\midrule
     % & Cactus   &0.0 &0.0 &0.0 &0.0 \\
    % Small Motion & ParkScene &0.0 &0.0 &0.0 &0.0 \\
    Small Motion & Johnny  &0.0 &78.5 &19.2 &\textbf{0.0} \\
    & FourPeople   &0.0 &41.2 &14.0 &\textbf{0.0} \\
    \bottomrule 
\end{tabular}
}
\vspace{-0.2cm}
\end{table}

\begin{table}[!t]
\centering
\renewcommand{\arraystretch}{1.15}
\scriptsize
\vspace{-0.1cm}
\caption{Ablation study on adaptive motion prediction method.}\label{ablation_study_amp}
\vspace{-0.3cm}
\resizebox{0.98\textwidth}{!}{
% \fontsize{7}{10}\selectfont%设置字体大小 %\midrule
% Accumulation-based Estimation (/wo OFR)
% Accumulation-based Estimation (/w OFR)
\begin{tabular}{c|ccccccc}
    \toprule  
        % \makebox[0.06\textwidth][c]{Sequences} & \makebox[0.06\textwidth][c]{Deirect} & \makebox[0.18\textwidth][c]{Accumulation\,(/wo OFR)} & \makebox[0.18\textwidth][c]{Accumulation\,(/w OFR)} & \makebox[0.06\textwidth][c]{Adaptive}\\
        \multirow{2}{*}{\makebox[0.05\textwidth][c]{Sequences}} & \multirow{2}{*}{\makebox[0.08\textwidth][c]{w/o MP}} & \makebox[0.1\textwidth][c]{Down Factor} & 
        \makebox[0.1\textwidth][c]{Down Factor} & 
        \makebox[0.1\textwidth][c]{Down Factor} & 
        \makebox[0.1\textwidth][c]{Down Factor} &
        \makebox[0.1\textwidth][c]{Down Factor} &
        \multirow{2}{*}{\makebox[0.02\textwidth][c]{AMP}}\\
        & & \{1\}& \{1,2\} & \{1,2,4\} & \{1,2,4,8\} & \{1,2,4,8,16\} &\\
        
    \hline  % &\textbf{\makecell{Flow-based warping \\ at all scale}} \makebox[0.05\textwidth][c]
    BasketballDrive   &0.0 &-8.2 &-9.2 &-9.8 &-10.2 &-10.2 &\textbf{-10.7} \\
    % Kimono   &0.0 &0.0 &0.0 &0.0 &0.0 &0.0 \\
    Jockey   &0.0 &-17.5 &-24.8 &-25.5 &-26.1 &-26.1 &\textbf{-26.6} \\ %\midrule
    % ReadySteadyGo   &0.0 &0.0 &0.0 &0.0 &0.0 &0.0 \\ %\midrule
    % YatchRide   &0.0 &0.0 &0.0 &0.0 &0.0 &0.0 \\ %\midrule
    % \hline
    HoneyBee   &0.0 &-4.8 &-4.8 &-4.8 &-4.8 &-4.8 &\textbf{-4.8} \\ %\midrule\midrule
     % & Cactus   &0.0 &0.0 &0.0 &0.0 \\
    % Small Motion & ParkScene &0.0 &0.0 &0.0 &0.0 \\
    % Johnny  &0.0 &0.0 &0.0 &0.0 &0.0 &0.0 \\
    % FourPeople   &0.0 &0.0 &0.0 &0.0 &0.0 &0.0 \\
    \bottomrule 
\end{tabular}
}
\vspace{-0.2cm}
\end{table}

\begin{table}[!t]
\renewcommand{\arraystretch}{1.2}
\scriptsize
\centering
\vspace{-0.2cm}
\caption{Model complexity comparison of learned methods.}\label{complexity_comparison}
\vspace{-0.3cm}
\resizebox{0.7\textwidth}{!}{
% \fontsize{7}{11}\selectfont%设置字体大小 %\midrule
\begin{tabular}{c|ccccc}
    \toprule  
     \makebox[0.1\textwidth][c]{Methods}& \makebox[0.08\textwidth][c]{Parameters} & \makebox[0.12\textwidth][c]{Encoding Time} & \makebox[0.12\textwidth][c]{Decoding Time} \\
    \hline  % &\textbf{\makecell{Flow-based warping \\ at all scale}} \makebox[0.05\textwidth][c]
    DCVC-DC\,(LD)  &19.78M &0.81s &0.61s \\ 
    % DCVC-FM\,(LD)   &M &576ms &436ms \\ %\midrule
    B-CANF\,(RA)  &46.77M &1.52s &0.98s \\ 
    % TLZMC**\,(RA)   &50.68M &576ms &436ms \\ %\midrule
    Ours\,(RA)  &22.47M &1.79s &0.80s \\ 
    % HEM*  &16.80 &6.92 &3.45 & & \\ 
    % HLGC(HEM*)   &21.48 &8.84 &4.42 & & \\ %\midrule
    \bottomrule 
\end{tabular}
}
\vspace{-0.3cm}
\end{table}

\noindent
\textbf{Adaptive Motion Estimation.} To analyze the design of our AME module, we conduct ablation studies on sequences of different motion types. The direct estimation method is set as anchor. As illustrated in Tab~\ref{ablation_study_ame}, the accumulation-based estimation method brings significant performance gain in large motion scenes but suffers performance loss in small motion scenes. Adding the optical flow refinement (OFR) module to the accumulation-based method brings obvious performance improvement. This is because the OFR module can effectively rectify the accumulation error. Comparison results show that our AME module can handle different types of motion.

\noindent
\textbf{Adaptive Motion Prediction.} We conduct ablation studies in Tab~\ref{ablation_study_amp} to analyze the design of our AMP module. The method (w/o MP) that does not adopt motion prediction is set as anchor. We gradually expand the downsampling factor set of the reference frame from \{1\} to \{1,2,4,8,16\}. We find that choosing larger downsampling factors helps alleviate the domain shift problem in large motion scenes \textit{BasketballDrive} and \textit{Jockey}. Our AMP does not use the motion prediction results when the motion prediction accuracy is poor, which is also proven to be effective. In small motion scene \textit{HoneyBee}, since the motion prediction network can handle small motions well, no gain is achieved when choosing larger downsampling factors.

\SubSection{Model Complexity}
\label{sec:computational_complexity}
In Tab~\ref{complexity_comparison}, we compare the model complexity with SOTA learned methods DCVC-DC (LD) and B-CANF (RA). All the models are run on a NVIDIA V100 GPU. For the encoding and decoding time, we report the model inference time with a 1080p frame as input. Compared with the P-frame coding method DCVC-DC, the model complexity of our L-LBVC is increased. This is mainly because the bi-directional motion and bi-directional context increase the complexity on both sides of the encoder and decoder. As for the comparison with the B-frame coding method B-CANF, our L-LBVC has better RD performance, fewer parameters, and faster decoding speed.
% In Tab~\ref{complexity_comparison}, we compare the model complexity with SOTA LBVC method B-CANF. All the models are run on a NVIDIA V100 GPU. For the encoding and decoding time, we report the model inference time with a 1080p frame as input. Comparison with B-frame coding method B-CANF, our L-LBVC has better RD performance, fewer parameters, and faster encoding/decoding speed.

% In Table~\ref{complexity_comparison}, we compare the model complexity in parameters, FLOPs, MACs, encoding time and decoding time with basline method TCM\cite{sheng2022temporal}. The experiment is conducted on NVIDIA GeForce RTX 3090 GPU. We use one 1080p frame to measure complexity. For the encoding and decoding time, we report the model inference time on GPU. Comparing with baseline, our HLGC method slightly increases the model complexity (4.5\% extra parameters). Our encoding and decoding time is increased a little. However, in terms of PSNR, our HLGC method brings 16.1\% bitrate saving over the strong basline TCM\cite{sheng2022temporal}. We think this is a cost worth paying.
% \Section{Conclusion}
\Section{Conclusion}
In this paper, we propose a novel LBVC framework named L-LBVC. To improve the accuracy of long-term motion estimation, we propose an adaptive motion estimation (AME) method that can effectively handle various motion ranges. The AME module applies direct estimation for adjacent frames and non-adjacent frames with small motions and applies accumulation-based estimation for long-term motions between non-adjacent frames. To improve the accuracy of long-term motion prediction, we propose an adaptive motion prediction (AMP) method that can largely reduce the bit cost for motion coding. The AMP module adapts the motion range by adaptively downsampling the reference frames for flow prediction. Experimental results showed that our L-LBVC can significantly outperform the previous SOTA LVC methods.

% In this paper, we propose a Hybrid Local-Global Context learning method to better generate high-quality contexts for neural video compression. For hybrid context generation, we combine the advantages of flow-based warping and deformable compensation in an optimal way. Our proposed method achieves better RD trade-off between compensation accuracy and bit cost for motion coding. Moreover, we design a local-global context enhancement module to further enhance the quality of contexts, which fully explore the local-global information of previous reconstructed signals. Experimental results on standard test datasets showed that our proposed HLGC method can significantly enhance the state-of-the-art methods.
\Section{Acknowledgment}
\sloppy{}
This work is financially supported by Guangdong Provincial Key Laboratory of Ultra High Definition Immersive Media Technology and National Natural Science Foundation of China U21B2012, this work is also financially supported for Outstanding Talents Training Fund in Shenzhen and R24115SG MIGU-PKU META VISION TECHNOLOGY INNOVATION LAB.

\begin{figure}[!h]
  \centering
  % \vspace{-0.2cm}
  \includegraphics[width=\linewidth, trim = {10 0 10 10}, clip]{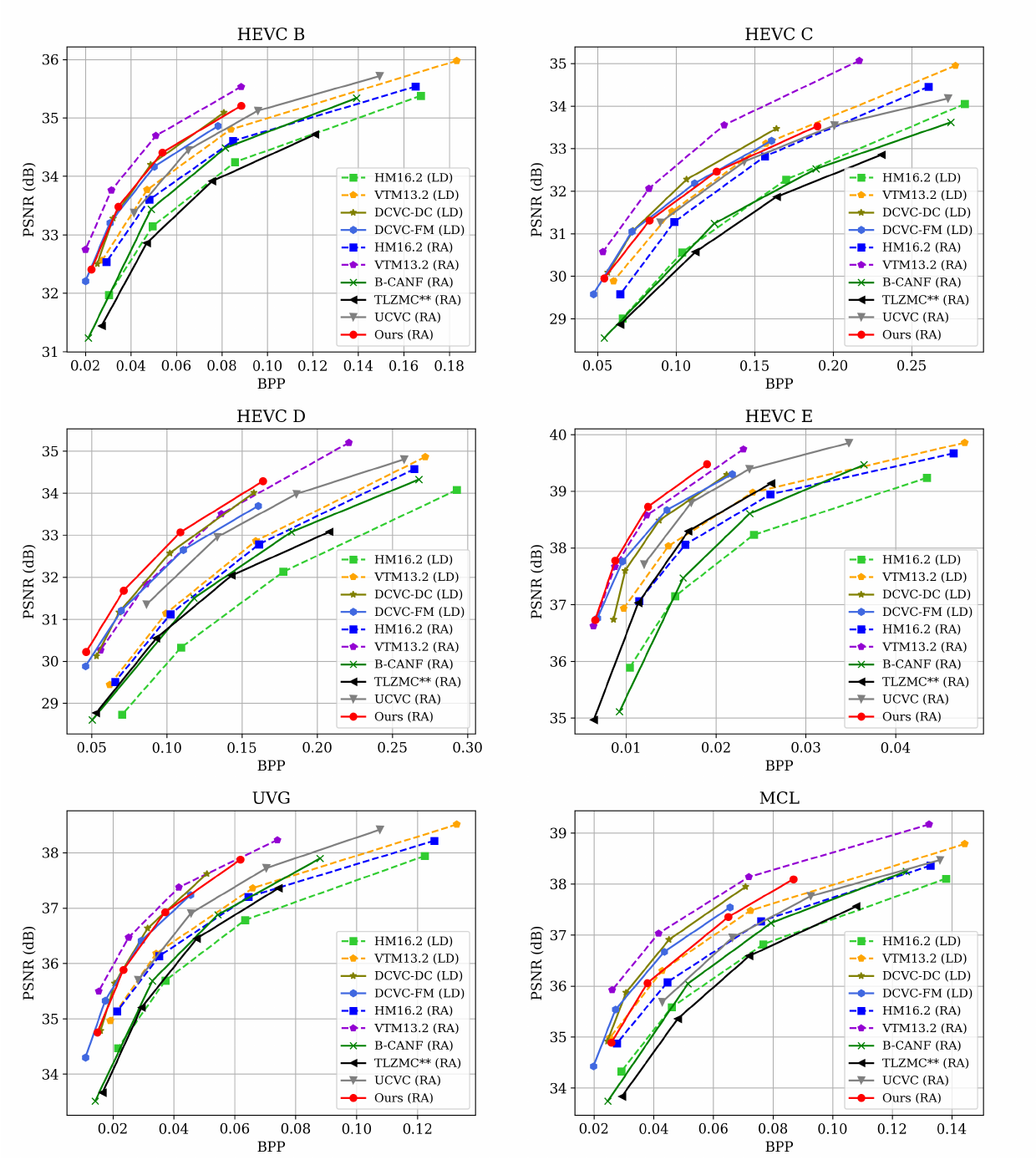}
  \vspace{-1.3cm}
  \caption{RD-curves on all datasets.}
  \label{rd_curves_llbvc}
  \vspace{-0.5cm}
\end{figure}

\Section{References}
\bibliographystyle{IEEEbib}
\bibliography{refs}
\Section{Appendix}
\SubSection{More RD Performance Results}
We draw the RD-curves of our L-LBVC and other methods on all datasets in Fig.~\ref{rd_curves_llbvc}. We set the intra period to 32 for all schemes and test 96 frames for each video. In Tab~\ref{rd_data_llbvc}, Tab~\ref{rd_data_bcanf}, Tab~\ref{rd_data_tlzmc} and Tab~\ref{rd_data_ucvc}, we also give the detailed RD data of L-LBVC and other LBVC methods on all datasets.

\begin{table}[!t]
\renewcommand{\arraystretch}{1.3}
\scriptsize
\centering
\vspace{-0.2cm}
\caption{Detailed RD data of L-LBVC on all datasets.}\label{rd_data_llbvc}
\vspace{-0.3cm}
\resizebox{0.95\textwidth}{!}{
% \fontsize{7}{11}\selectfont%设置字体大小 %\midrule
\begin{tabular}{c|ccccc}
    \toprule  
     \makebox[0.2\textwidth][c]{Datasets}& \makebox[0.2\textwidth][c]{BPP} & \makebox[0.2\textwidth][c]{PSNR}  \\
    \hline  % &\textbf{\makecell{Flow-based warping \\ at all scale}} \makebox[0.05\textwidth][c]
      HEVC B  &0.022541 &32.4048936  \\ 
      &0.0343894 &33.4788728  \\ 
      &0.0537992 &34.4085072  \\
      &0.0884698 &35.2068996  \\ 
      \hline
      HEVC C  &0.054325 &29.952718  \\ 
      &0.0831585 &31.309173  \\ 
      &0.125748 &32.45687725  \\
      &0.18984875 &33.52555425  \\ 
      \hline
      HEVC D  &0.04630625 &30.21744675  \\ 
      &0.071234 &31.67974325  \\ 
      &0.1089715 &33.068674  \\
      &0.16387575 &34.284673  \\ 
      \hline
      HEVC E  &0.006540333 &36.72886733  \\ 
      &0.008792 &37.774594  \\ 
      &0.012414667 &38.72468333  \\
      &0.019001333 &39.477422  \\ 
      \hline
      UVG  &0.014937857 &34.75299  \\ 
      &0.023431429 &35.88392657  \\ 
      &0.03715 &36.924816  \\
      &0.061856286 &37.87352014  \\
      \hline
      MCL-JCV  &0.0258331 &34.89034223  \\ 
      &0.037863233 &36.056121  \\ 
      &0.0650062 &37.3533327  \\
      &0.0868334 &38.09305313  \\ 
    \bottomrule 
\end{tabular}
}
\vspace{-0.3cm}
\end{table}

\begin{table}[!t]
\renewcommand{\arraystretch}{1.2}
\scriptsize
\centering
\vspace{-0.2cm}
\caption{Detailed RD data of B-CANF on all datasets.}\label{rd_data_bcanf}
\vspace{-0.3cm}
\resizebox{0.95\textwidth}{!}{
% \fontsize{7}{11}\selectfont%设置字体大小 %\midrule
\begin{tabular}{c|ccccc}
    \toprule  
     \makebox[0.2\textwidth][c]{Datasets}& \makebox[0.2\textwidth][c]{BPP} & \makebox[0.2\textwidth][c]{PSNR}  \\
    \hline  % &\textbf{\makecell{Flow-based warping \\ at all scale}} \makebox[0.05\textwidth][c]
      HEVC B  &0.021258263 &31.23296644  \\ 
      &0.048835651 &33.43850815  \\ 
      &0.081538287 &34.48546895  \\
      &0.139200355 &35.33930919  \\ 
      \hline
      HEVC C  &0.054342171 &28.54487321  \\ 
      &0.124310023 &31.2350908  \\ 
      &0.189061085 &32.52133016  \\
      &0.274959218 &33.61745382  \\ 
      \hline
      HEVC D  &0.050400371 &28.60126077  \\ 
      &0.118140821 &31.50454822  \\ 
      &0.182832898 &33.07577646  \\
      &0.267706819 &34.32540801  \\ 
      \hline
      HEVC E  &0.00924396 &35.10945335  \\ 
      &0.016339867 &37.46816448  \\ 
      &0.023757079 &38.60589943  \\
      &0.036484746 &39.46747866  \\ 
      \hline
      UVG  &0.014185229 &33.51034995  \\ 
      &0.032932138 &35.68262773  \\ 
      &0.054444912 &36.87874378  \\
      &0.088018877 &37.89475445  \\ 
      \hline
      MCL-JCV  &0.024603117 &33.74068932  \\ 
      &0.051520658 &36.04693306  \\ 
      &0.079336038 &37.23407954  \\ 
      &0.124722931 &38.25016683  \\
    \bottomrule 
\end{tabular}
}
\vspace{-0.3cm}
\end{table}

\begin{table}[!t]
\renewcommand{\arraystretch}{1.2}
\scriptsize
\centering
\vspace{-0.2cm}
\caption{Detailed RD data of TLZMC** on all datasets.}\label{rd_data_tlzmc}
\vspace{-0.3cm}
\resizebox{0.98\textwidth}{!}{
% \fontsize{7}{11}\selectfont%设置字体大小 %\midrule
\begin{tabular}{c|ccccc}
    \toprule  
     \makebox[0.2\textwidth][c]{Datasets}& \makebox[0.2\textwidth][c]{BPP} & \makebox[0.2\textwidth][c]{PSNR}  \\
    \hline  % &\textbf{\makecell{Flow-based warping \\ at all scale}} \makebox[0.05\textwidth][c]
      HEVC B  &0.027100833 &31.44163333  \\ 
      &0.04696375 &32.85978125  \\ 
      &0.075663125 &33.91563125  \\
      &0.121033125 &34.71725208  \\ 
      \hline
      HEVC C  &0.06432526 &28.86123438  \\ 
      &0.112341146 &30.56802865  \\ 
      &0.163988542 &31.86539844  \\
      &0.230882813 &32.8546849  \\ 
      \hline
      HEVC D  &0.052945833 &28.7648724  \\ 
      &0.09295 &30.54558073  \\ 
      &0.143082813 &32.03923958  \\
      &0.208146615 &33.07934896  \\ 
      \hline
      HEVC E  &0.006390972 &34.96432986  \\ 
      &0.011330903 &37.02615972  \\ 
      &0.016922917 &38.29631597  \\
      &0.026126736 &39.13963889  \\ 
      \hline
      UVG  &0.016638988 &33.66485268  \\ 
      &0.029423065 &35.2036369  \\ 
      &0.04754628 &36.44982589  \\
      &0.074393155 &37.36014583  \\ 
      \hline
      MCL-JCV  &0.029364444 &33.83543611  \\ 
      &0.048047847 &35.35591285  \\ 
      &0.072036146 &36.5876  \\
      &0.10796059 &37.56282396  \\  
    \bottomrule 
\end{tabular}
}
\vspace{-0.3cm}
\end{table}

\begin{table}[!t]
\renewcommand{\arraystretch}{1.2}
\scriptsize
\centering
\vspace{-0.2cm}
\caption{Detailed RD data of UCVC on all datasets.}\label{rd_data_ucvc}
\vspace{-0.3cm}
\resizebox{0.95\textwidth}{!}{
% \fontsize{7}{11}\selectfont%设置字体大小 %\midrule
\begin{tabular}{c|ccccc}
    \toprule  
     \makebox[0.2\textwidth][c]{Datasets}& \makebox[0.2\textwidth][c]{BPP} & \makebox[0.2\textwidth][c]{PSNR}  \\
    \hline  % &\textbf{\makecell{Flow-based warping \\ at all scale}} \makebox[0.05\textwidth][c]
      HEVC B  &0.04131938 &33.37425194  \\ 
      &0.065140432 &34.45109787  \\ 
      &0.095882587 &35.12384872  \\
      &0.14938307 &35.71579514  \\ 
      \hline
      HEVC C  &0.089994 &31.26379  \\ 
      &0.143457 &32.69326  \\ 
      &0.200778 &33.5438  \\
      &0.273218 &34.17773  \\ 
      \hline
      HEVC D  &0.086545974 &31.35397577  \\ 
      &0.133526142 &32.95447254  \\ 
      &0.186219618 &33.98046207  \\
      &0.25767478 &34.802351  \\ 
      \hline
      HEVC E  &0.011984954 &37.70845413  \\ 
      &0.017255015 &38.80128225  \\ 
      &0.023733483 &39.39171855  \\
      &0.034794078 &39.85059611  \\ 
      \hline
      UVG  &0.028256288 &35.69674138  \\ 
      &0.045621969 &36.9057263  \\ 
      &0.070355627 &37.72010149  \\
      &0.107674897 &38.41659219  \\ 
      \hline
      MCL-JCV  &0.04286498 &35.68534851  \\ 
      &0.066373468 &36.94601498  \\ 
      &0.092703227 &37.76239993  \\
      &0.136028244 &38.46938661  \\  
    \bottomrule 
\end{tabular}
}
\vspace{-0.3cm}
\end{table}
\end{document}